\documentclass[10pt,twocolumn,letterpaper]{article}

\usepackage[final]{cvpr}      
\usepackage[accsupp]{axessibility}  
\usepackage{times}
\usepackage{epsfig}
\usepackage{graphicx}
\usepackage{amsmath}
\usepackage{amssymb}
\usepackage{algorithm}
\usepackage{algorithmic}
\usepackage{listings}
\usepackage{soul}
\usepackage{cancel}
\usepackage{dsfont}
\usepackage{pifont}
\usepackage[utf8]{inputenc}
\usepackage{multirow}
\usepackage[numbers,sort,compress]{natbib}
\usepackage[usenames,dvipsnames,svgnames,table]{xcolor}
\usepackage{tablefootnote} 
\graphicspath{{./}{./images/}}
\usepackage{subcaption}
\usepackage{xspace}
\usepackage{tabularx,ragged2e,booktabs}
\usepackage{color}
\definecolor{_fbteal3}{HTML}{CBCBCB} 
\definecolor{Gray}{gray}{0.85}
\definecolor{darkgreen}{rgb}{0.0, 0.5, 0.0}
\definecolor{darkred}{rgb}{0.64, 0.0, 0.0}
\newcolumntype{a}{>{\columncolor{_fbteal3}}c}
\usepackage[pagebackref=true,breaklinks=true,colorlinks,bookmarks=false]{hyperref}

\def\cvprPaperID{4075} 
\def\confName{CVPR}
\def\confYear{2023}

\makeatletter
\DeclareRobustCommand\onedot{\futurelet\@let@token\@onedot}
\def\@onedot{\ifx\@let@token.\else.\null\fi\xspace}

\def\eg{\textit{e.g}\onedot} 
\def\ie{\textit{i.e}\onedot} 
 
\def\etc{\emph{etc}\onedot} 

\def\etal{{et al}\onedot}
\makeatother

\def\Vec#1{{\boldsymbol{#1}}}

\newcommand{\cmark}{\ding{51}}
\newcommand{\xmark}{\ding{55}}
\newcommand{\putouralg}{\textsc{ProtoCon}\xspace}
\newcommand{\vx}{\boldsymbol{x}}
\newcommand{\vy}{\boldsymbol{y}}
\newcommand{\vz}{\boldsymbol{z}}
\newcommand{\vu}{\boldsymbol{u}}
\newcommand{\vq}{\boldsymbol{q}}
\newcommand{\vp}{\boldsymbol{p}}

\newcommand{\Lcal}{\mathcal{L}}
\newcommand{\Qcal}{\mathcal{Q}}
\newcommand{\Ical}{\mathcal{I}}
\newcommand{\Ocal}{\mathcal{O}}
\newcommand{\Xcal}{\mathcal{X}}
\newcommand{\Ucal}{\mathcal{U}}
\newcommand{\Acal}{\mathcal{A}}

\newcommand{\Pcal}{\mathcal{P}}

\DeclareMathOperator*{\argmax}{arg\,max}

\title{\vspace{-1.2em} \putouralg: Pseudo-label Refinement via Online Clustering and \underline{Proto}typical \underline{Con}sistency for Efficient Semi-supervised Learning}

\author{Islam Nassar\textsuperscript{1}
\quad Munawar Hayat\textsuperscript{1} 
\quad Ehsan Abbasnejad\textsuperscript{2} 
\quad Hamid Rezatofighi\textsuperscript{1} \\ 
Gholamreza Haffari\textsuperscript{1}\\
{\small \textsuperscript{1} Data Science and AI Department, Monash University, Australia – firstname.lastname@monash.edu} \\
{\small \textsuperscript{2} Australian Institute for Machine Learning, The University of Adelaide, Australia – firstname.lastname@adelaide.edu.au}
}

\date{October 2020}

\begin{document}
\maketitle    

\begin{abstract}
Confidence-based pseudo-labeling is among the dominant approaches in semi-supervised learning (SSL). It relies on including high-confidence predictions made on unlabeled data as additional targets to train the model. We propose \putouralg, a novel SSL method aimed at the less-explored label-scarce SSL where such methods usually underperform. \putouralg refines the pseudo-labels by leveraging their nearest neighbours' information. The neighbours are identified as the training proceeds using an online clustering approach operating in an embedding space trained via a prototypical loss to encourage well-formed clusters. The online nature of \putouralg allows it to utilise the label history of the entire dataset in one training cycle to refine labels in the following cycle without the need to store image embeddings. Hence, it can seamlessly scale to larger datasets at a low cost. Finally, \putouralg addresses the poor training signal in the initial phase of training (due to fewer confident predictions) by introducing an auxiliary self-supervised loss.  It delivers significant gains and faster convergence over state-of-the-art across 5 datasets, including CIFARs, ImageNet and DomainNet. 


\end{abstract}

\section{Introduction}
\label{sec:introduction}

\begin{figure}[!t]
 \centering
  \includegraphics[width=0.47\textwidth]{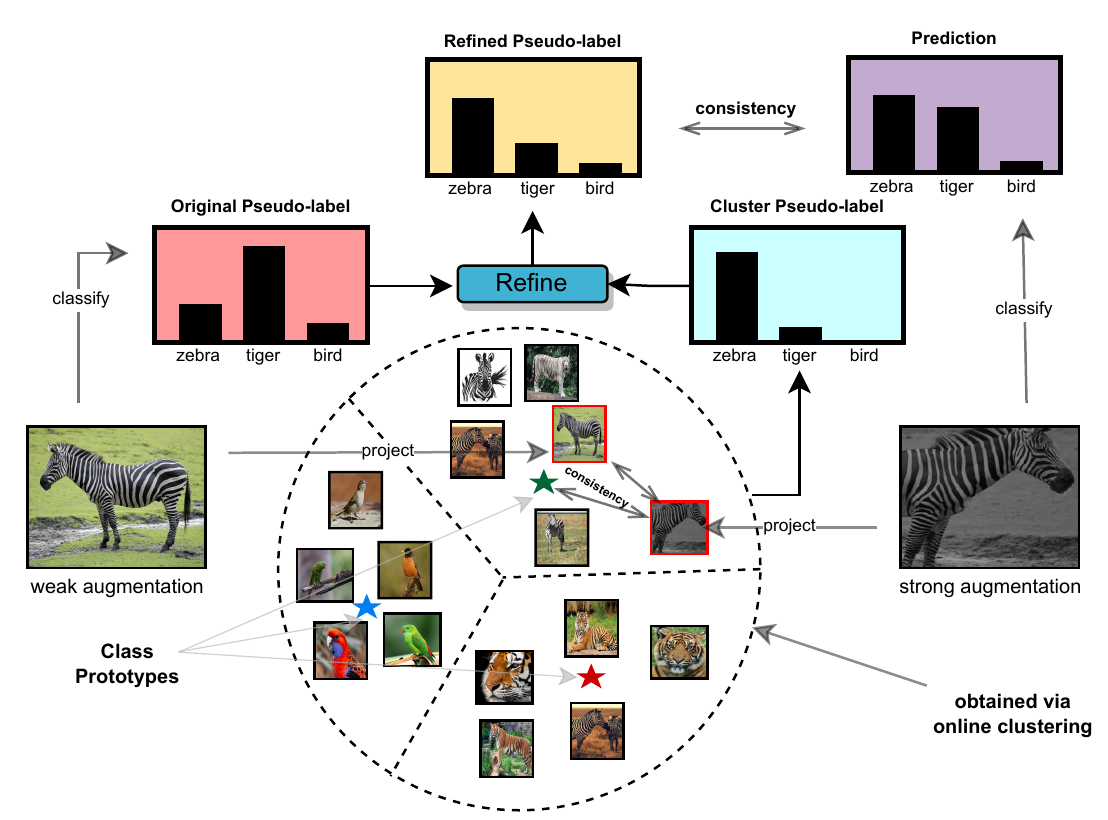}
 \caption{\putouralg refines a pseudo-label of a given sample by knowledge of its neighbours in a prototypical embedding space. Neighbours are identified in an online manner using constrained K-means clustering. Best viewed zoomed in.}
 \vspace{-5mm}
 \label{fig:high_level_diagram}
\end{figure}

Semi-supervised Learning (SSL)~\cite{van2020survey, chapelle2009semi} leverages unlabeled data to guide learning from a small amount of labeled data; thereby, providing a promising alternative to costly human annotations.
In recent years, SSL frontiers have seen substantial advances through confidence-based pseudo-labeling~\cite{lee2013_pseudo, Sohn_fixmatch20, xie2019_uda, li2021comatch, wang2022debiased}. In these methods, a model iteratively generates pseudo-labels for unlabeled samples which are then used as targets to train the model. To overcome confirmation bias~\cite{arazo_pseudo, nassar2021all} \ie, the model being biased by training on its own wrong predictions, these methods only retain samples with high confidence predictions for pseudo-labeling; thus ensuring that only reliable samples are used to train the model. While confidence works well in moderately labeled data regimes, it usually struggles in label-scarce settings\footnote{\scriptsize We denote settings with less than 10 images per class as ``label-scarce.''}. This is primarily because the model becomes over-confident about the more distinguishable classes~\cite{nguyen2015deep, hein2019relu} faster than others, leading to a collapse.

In this work, we propose \putouralg, a novel method which addresses such a limitation in label-scarce SSL. Its key idea is to complement confidence with a label refinement strategy to encourage more accurate pseudo-labels. 
To that end, we perform the refinement by adopting a co-training~\cite{blum1998_co_training} framework: for each image, we obtain two different labels and combine them to obtain our final pseudo-label. The first is the model's softmax prediction, whereas the second is an aggregate pseudo-label describing the image's neighbourhood based on the pseudo-labels of other images in its vicinity. However, a key requirement for the success of co-training is to ensure that the two labels are obtained using sufficiently different image representations~\cite{van2020survey} to allow the model to learn based on their disagreements. As such, we employ a non-linear projection to map our encoder's representation into a different embedding space. We train this projector jointly with the model with a prototypical consistency objective to ensure it learns a different, yet relevant, mapping for our images. Then we define the neighbourhood pseudo-label based on the vicinity in that embedding space. In essence, we minimise a sample bias by smoothing its pseudo-label in class space via knowledge of its neighbours in the prototypical space.

Additionally, we design our method to be fully online, enabling us to scale to large datasets at a low cost.
We identify neighbours in the embedding space on-the-fly as the training proceeds by leveraging online K-means clustering. This alleviates the need to store expensive image embeddings~\cite{li2021comatch}, or to utilise offline nearest neighbour retrieval~\cite{li2021learning, zheng2022simmatch}. However, applying naive K-means risks collapsing to only a few imbalanced clusters making it less useful for our purpose. Hence, we employ a constrained objective~\cite{bradley2000constrained} lower bounding each cluster size; thereby, ensuring that each sample has enough neighbours in its cluster. We show that the online nature of our method allows it to leverage the entire prediction history in one epoch to refine labels in the subsequent epoch at a fraction of the cost required by other methods and with a better performance.

\putouralg's final ingredient addresses another limitation of confidence-based methods: since the model only retains high confident samples for pseudo-labeling, the initial phase of the training usually suffers from a weak training signal due to fewer confident predictions. In effect, this leads to only learning from the very few labeled samples which destabilises the training potentially due to overfitting~\cite{lucas2022barely}. To boost the initial training signal, we adopt a self-supervised instance-consistency~\cite{grill2020bootstrap, caron2021emerging} loss applied on samples that fall below the threshold. Our choice of loss is more consistent with the classification task as opposed to contrastive instance discrimination losses~\cite{he2020momentum, chen2020simple} which treat each image as its own class. This helps our method to converge faster without loss of accuracy.

\begin{figure*}[h!]
 \centering
  \scalebox{0.9}{\includegraphics[width=0.97\textwidth]{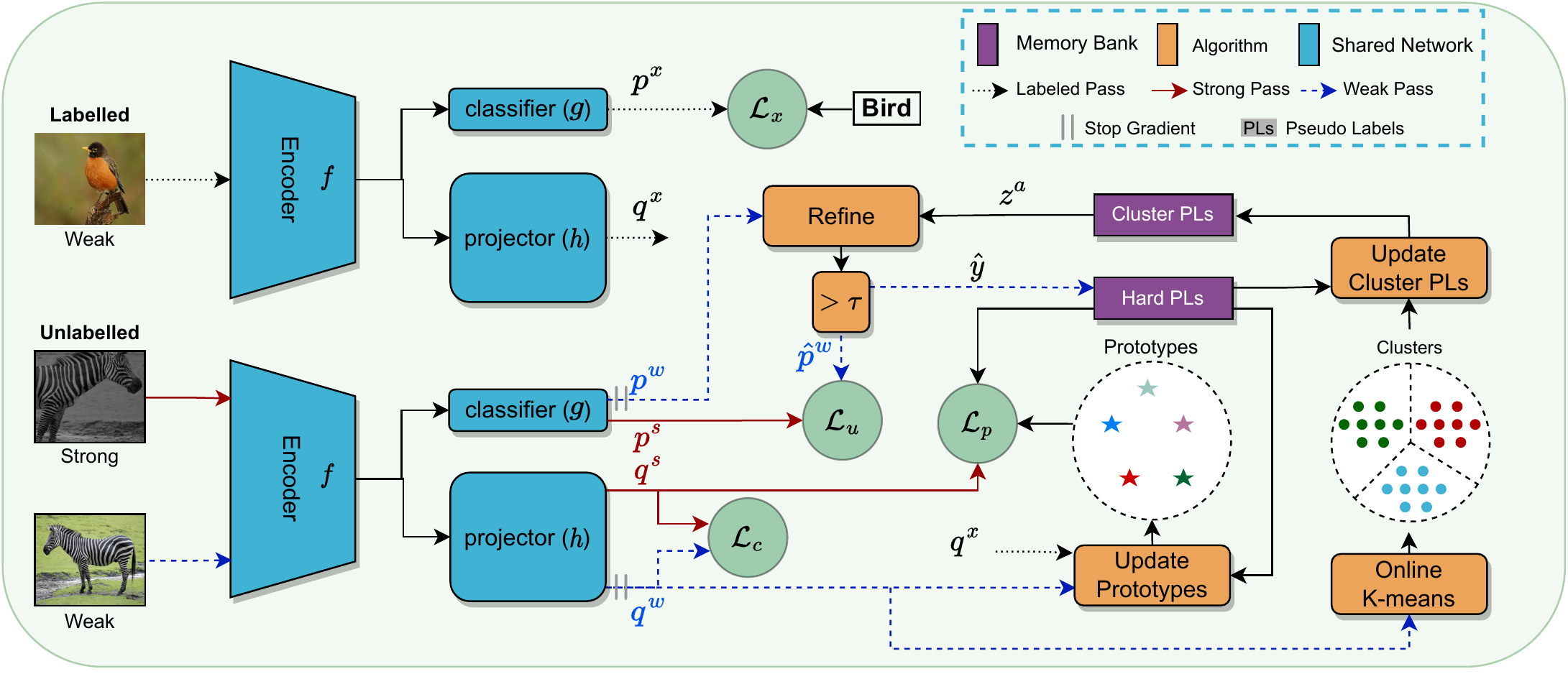}}
 \caption{\textbf{Method overview.} A soft pseudo-label $p^w$ is first obtained based on the weak view. Then it is refined using the sample's cluster pseudo-label $z^a$ before using it as target in $\Lcal_u$. Clustering assignments $a$ are calculated online using the projections of the weak samples $q^w$ in the embedding space $h$ which is trained via a prototypical loss $\Lcal_p$. Prototype targets are updated once after each epoch by averaging the accumulated projections of reliable samples for each class throughout the epoch. Cluster pseudo-labels are updated after each epoch using the cluster assignments/scores of all the samples and their respective hard pseudo-labels $\hat{y}$. Finally, the self-supervised loss $\Lcal_c$ ensures consistency between projections $q^s$ and $q^w$.}
 \label{fig:main_protocon}
 \vspace{-5mm}
\end{figure*}

We demonstrate \putouralg's superior performance against comparable state-of-the-art methods on 5 datasets including CIFAR, ImageNet and DomainNet. Notably, \putouralg achieves 2.2\%, 1\% improvement on the SSL ImageNet protocol with 0.2\% and 1\% of the labeled data, respectively. Additionally, we show that our method exhibits faster convergence and more stable initial training compared to baselines, thanks to our additional self-supervised loss. In summary, our contributions are:
\begin{itemize}
\setlength{\itemsep}{1pt}
\setlength{\parsep}{1pt}
\setlength{\parskip}{0.5pt}
    \item We propose a memory-efficient method addressing confirmation bias in label-scarce SSL via a novel label refinement strategy based on co-training.
    \item We improve training dynamics and convergence of confidence-based methods by adopting self-supervised losses to the SSL objective.
    \item We show state-of-the-art results on 5 SSL benchmarks.
\end{itemize}
\section{Background}
\label{sec:background}

We begin by reviewing existing SSL approaches with a special focus on relevant methods in the low-label regime. 

\smallskip
\noindent \textbf{Confidence-based pseudo-labeling} is an integral component in most of recent SSL methods~\cite{Sohn_fixmatch20, li2021comatch, wang2022debiased, nassar2021all, kuo2020featmatch}. However, recent research shows that using a fixed threshold underperforms in low-data settings because the model collapses to the few easy-to-learn classes early in the training. Some researchers combat this effect by using class-~\cite{zhang2021flexmatch} or instance-based~\cite{xu2021dash} adaptive thresholds, or by aligning~\cite{Berthelot_RemixMatch19} or debiasing~\cite{wang2022debiased} the pseudo-label distribution by keeping a running average of pseudo-labels to avoid the inherent imbalance in pseudo-labels. Another direction focuses on pseudo-label refinement, whereby the classifier's predictions are adjusted by training another projection head on an auxiliary task such as weak-supervision via language semantics~\cite{nassar2021all}, instance-similarity matching~\cite{zheng2022simmatch}, or graph-based contrastive learning~\cite{li2021comatch}. Our method follows the refinement approach, where we employ online constrained clustering to leverage nearest neighbours information for refinement. Different from previous methods, our method is fully online and hence allows using the entire prediction history in one training epoch to refine pseudo-labels in the subsequent epoch with minimal memory requirements.    

\smallskip
\noindent \textbf{Consistency Regularization} combined with pseudo-labeling underpins many recent state-of-the-art SSL methods~\cite{xie2019_uda, li2019decoupled, Sajjadi_NIPS16, Berthelot_MixMatch19, Sohn_fixmatch20, li2021comatch, kuo2020featmatch}; it exploits the smoothness assumption~\cite{van2020survey} where the model is expected to produce similar pseudo-labels for minor input perturbations. The seminal FixMatch~\cite{Sohn_fixmatch20} and following work~\cite{nassar2021all, li2021comatch, wang2022debiased} leverage this idea by obtaining pseudo-labels through a weak form of augmentation and applying the loss against the model's prediction for a strong augmentation. Our method utilises a similar approach, but different from previous work, we additionally apply an instance-consistency loss in our projection embedding space.


\smallskip
\noindent \textbf{Semi-supervision via self-supervision} is gaining recent popularity due to the incredible success of self-supervised learning for model pretraining. Two common approaches are: 1) performing self-supervised pretraining followed by supervised fine-tuning on the few labeled samples~\cite{chen2020simple, chen2020big, grill2020bootstrap, nassar2022lava, caron2021emerging}, and 2) including a self-supervised loss to the semi-supervised objective to enhance training~\cite{zhai2019s4l, lucas2022barely, wallin2022doublematch, li2021comatch, zheng2022simmatch}. However, the choice of the task is crucial: tasks such as instance discrimination~\cite{he2020momentum, chen2020simple}, which treats each image as its own class, can hurt semi-supervised image classification as it partially conflicts with it. Instead, we use an instance-consistency loss akin to that of~\cite{caron2021emerging} to boost the initial training signal by leveraging samples which are not retained for pseudo-labeling in the early phase of the training.

\section{\putouralg}
\label{sec:method}
\noindent \textbf{Preliminaries.} We consider a semi-supervised image classification problem, where we train a model using $M$ labeled samples and $N$ unlabeled samples, where $N>>M$. We use mini-batches of labeled instances, $\mathcal{X} = \{(\vx_j, \vy_j)\}_{j=1}^B$ and unlabeled instances, $\mathcal{U} = \{\Vec{u}_i\}_{i=1}^{\mu \cdot B}$, where the scalar $\mu$ denotes the ratio between the number of unlabeled and labeled examples in a given batch, and $\vy$ is the one-hot vector of the class label $c \in \{1, \dots, C\}$. We employ an encoder network $f$ to get latent representations $f(.)$. We attach a softmax classifier $g(\cdot)$, which produces a distribution over classes $\vp = g \circ f$. Moreover, we attach a projector  $h(\cdot)$, an MLP followed by an $\ell_2$ norm layer, to get a normalised embedding $\vq \in \mathbb{R}^{d} = h \circ f$. Following~\cite{Sohn_fixmatch20}, we apply weak augmentations $\Acal_w(\cdot)$ on all images and an additional strong augmentation~\cite{cubuk2020_randaugment} $\Acal_s(\cdot)$ only on unlabeled ones. 

\smallskip
\noindent \textbf{Motivation.} Our aim is to refine pseudo-labels before using them to train our model in order to minimise confirmation bias in label-scarce SSL. We achieve this via a co-training approach (see Fig.~\ref{fig:main_protocon}). For each image, we obtain two pseudo-labels and combine them to obtain our final pseudo-label $\hat{\vp}^w$. The first is the classifier softmax prediction ${\vp}^w$ based on a weakly augmented image, whereas the second is an aggregate pseudo-label $\vz^a$ describing the sample's neighbourhood. To ensure the two labels are based on sufficiently different representations, we define an image neighbourhood\footnote{\scriptsize We use ``neighbourhood'' and ``cluster'' interchangeably.} via online clustering in an embedding space obtained via projector $h$ and trained for prototypical consistency instead of class prediction.  Projector $h$ and classifier $g$ are jointly trained with the encoder $f$, while interacting over pseudo-labels. The classifier is trained with pseudo-labels which are refined based on their nearest neighbours in the embedding space, whereas the projector $h$ is trained using prototypes obtained based on the refined pseudo-labels to impose structure on the embedding space.

\smallskip
\noindent \textbf{Prototypical Space.}
Here, we discuss our procedure to learn our embedding space defined by $h$. Inspired by prototypical learning~\cite{snell2017prototypical}, we would like to encourage well-clustered image projections in our space by attracting samples to their class prototypes and away from others. Hence, we employ a contrastive objective using the class prototypes as targets rather than the class labels. We calculate class prototypes at the end of a given epoch by knowledge of the ``reliable'' samples in the previous epoch. Specifically, by employing a memory bank of $\Ocal(2N)$, we keep track of samples hard pseudo-labels $\{\hat{y}_i = \argmax(\hat{\vp}^w_i) \forall \vu_i \in \Ucal\}$ in a given epoch; as well as a reliability indicator for each sample $\eta_i = \mathds{1} (\max (\hat{\vp}^w_i) \geq \tau)$ denoting if its max prediction exceeds the confidence threshold $\tau$. Subsequently, we update the prototypes $\Pcal \in \mathbb{R}^{C \times d}$ as the average projections (accumulated over the epoch) of labeled images and reliable unlabeled images. Formally, let $\Ical^x_c = \{i | \forall \vx_i \in \Xcal, y_i=c \}$ be the indices of labelled instances with true class $c$, and $\Ical^w_c = \{i | \forall \vu_i \in \Ucal, \eta_i=1, \hat{y}_i=c \}$ be the indices of the reliable unlabelled samples with hard pseudo-label $c$. The normalised prototype for class $c$ can then be obtained as per:
\begin{equation}
	\bar{\Pcal}_c = \frac{\sum_{i \in \Ical^x_c \cup \Ical^w_c} \vq_i}{|\Ical^x_c| + |\Ical^w_c|} , \quad \Pcal_c = \frac{\bar{\Pcal}_c}{||\bar{\Pcal}_c||_2}
\end{equation}


\noindent Subsequently, in the following epoch, we minimize the following contrastive prototypical consistency loss on unlabeled samples:
\begin{equation}
    \Lcal_{p} = - \frac{1}{\mu B} \sum_{i=1}^{\mu B} \log \frac{\exp(\vq^s_i \cdot \Pcal_{\hat{y}_i} / T)}{\sum_{c=1}^{C} \exp(\vq^s_i \cdot \Pcal_c / T))}, 
\end{equation}
where $T$ is a temperature parameter. Note that the loss is applied against the projection of the strong augmentations to achieve consistency regularisation as in~\cite{Sohn_fixmatch20}.

\begin{algorithm}[t]
\caption{Pseudo-code of one epoch of \putouralg}
\label{alg:code}
\definecolor{codeblue}{rgb}{0.25,0.5,0.5}
\lstset{
  backgroundcolor=\color{white},
  basicstyle=\fontsize{7.2pt}{7.2pt}\ttfamily\selectfont,
  columns=fullflexible,
  breaklines=true,
  captionpos=b,
  commentstyle=\fontsize{7.2pt}{7.2pt}\color{codeblue},
  keywordstyle=\fontsize{7.2pt}{7.2pt},
}
\begin{lstlisting}[language=python]
# f, g, h: encoder, classifier, and projector
# b_x: labeled batch
# b_w, b_s: weak, strong unlabeled batches
# u_id: unique index of unlabeled samples
# N, C: num unlabeled samples, num classes
# CA: cluster assignment bank (N x 2)
# CPL: clusters pseudo-label bank (N x C)
# PH: samples pseudo-label bank (N x 1)
# Q: cluster centers
# P: class prototypes
# P_acc: prototypes accumulator
# alpha: pseudo-label refinement ratio

for b_x, b_w, b_s, u_id in loader:  
    # forward images and obtain p and q
    p_x, p_w, p_s = f(g(b_x, b_w, b_s))
    q_x, q_w, q_s = f(h(b_x, b_w, b_s))
    # calculate and save cluster assignment
    CA[u_id] = calc_clust_assignmnt(q_w, Q)  # Eqn.5
    # update centers and dual variables
    Q = update_cluster_centers(q_w)        # Eqn. 6 & 7
    #retrieve cluster pseudo-labels from previous epoch
    z = CPL[CA[u_id]]
    # refine p_w
    p_w_hat = alpha*p_w + (1 - alpha)*z   # Eqn. 9
    # save hard pseudo-labels
    PH[u_id] = argmax(p_w_hat)
    # accumulate prototypes (of reliable samples only)
    P_acc = accum_prototypes(q_x, q_w, PH[u_id])
    # apply losses (except in first epoch)
    Lx, Lp, Lu, Lc = backward_losses()  # Eqn. 2, 10-12
# after each epoch
P = update_prototypes(P_acc)  # Eqn. 1
CPL = calc_cluster_pseudo_labels(CA, PH) # Eqn. 8
\end{lstlisting}
\end{algorithm}

   

\smallskip
\noindent \textbf{Online Constrained K-means}
Here, the goal is to cluster instances in the prototypical space as a training epoch proceeds, so the cluster assignments (capturing the neighbourhood of each sample) are used to refine their pseudo-labels in the following epoch. We employ a mini-batch version of K-means~\cite{sculley2010web}. To avoid collapsing to one (or a few) imbalanced clusters, we ensure that each cluster has sufficient samples by enforcing a constraint on the lowest acceptable cluster size. Given our $N$ unlabeled projections, we cluster them into $K$ clusters defined by centroids $\Qcal = [\mathbf{c}_1, \cdots, \mathbf{c}_K] \in \mathbb{R}^{d \times K}$. We use the constrained K-means objective proposed by~\cite{bradley2000constrained}:
\begin{footnotesize}
\begin{align}
\label{eqn:ckmeans}
\min_{\Qcal, \mu \in \Delta} \sum_{i=1,k=1}^{i=N,k=K}\mu_{i,k}\|\vq_i - \mathbf{c}_k\|_2^2 \quad s.t.\quad \forall k\quad \sum_{i=1}^N \mu_{i,k}\geq \gamma
\end{align}
\end{footnotesize}

where $\gamma$ is the lower-bound of cluster size, $\mu_{i,k}$ is the assignment of the $i$-th unlabeled sample to the $k$-th cluster, and $\Delta=\{\mu|\forall i,\ \sum_{k}\mu_{i,k}=1,\forall i,k, \mu_{i,k}\in[0,1]\}$ is the domain of $\mu$. Subsequently, to solve Eqn.~\ref{eqn:ckmeans} in an online mini-batch manner, we adopt the alternate solver proposed in~\cite{qian2022unsupervised}. For a fixed $\Qcal$, the problem for updating $\mu$ can be simplified as an assignment problem. By introducing dual variables $\rho_k$ for each constraint $\sum_i \mu_{i,k} \geq \gamma$, the assignment can be obtained by solving the problem:
\begin{eqnarray}
\label{eqn:assignment}
\max_{\mu_{i}\in\Delta}  \sum_k s_{i,k} \mu_{i,k} + \sum_k\rho_k^{t-1} \mu_{i,k}
\end{eqnarray}
where $s_{i,k} = \vq_i^\top \mathbf{c}_k$ is the similarity between the projection of unlabeled sample $\vu_i$ and the $k$-th cluster centroid, and $t$ is the mini-batch iteration counter. Eqn.~\ref{eqn:assignment} can then be solved with the closed-form solution:
\begin{eqnarray}
\label{eq:solution}
\mu_{i,k} = \left\{\begin{array}{cc}1&k=\arg\max_k s_{i,k}+\rho_k^{t-1}\\0&o.w.\end{array}\right.
\end{eqnarray}
After assignment, dual variables are updated as\footnote{\scriptsize Refer to~\cite{qian2022unsupervised} and supplements for proofs of optimality and more details.} :
\begin{eqnarray}
\label{eq:finalrho}
\rho_k^t = \max\{0,\rho_k^{t-1} - \lambda \frac{1}{B}\sum_{i=1}^B( \mu_{i,k}^t-\frac{\gamma}{N})\}
\end{eqnarray}
where $\lambda$ is the dual learning rate.
Finally, we update the cluster centroids after each mini-batch\footnote{\scriptsize See supplements for a discussion about updating the centers every mini-batch opposed to every epoch.} as:
\begin{equation}
\label{eq:updatemulti}
\bar{\mathbf{c}_{k}}^t = \frac{\sum_i^m\mu_{i,k}^t\vq_i^t}{\sum_i^m \mu_{i,k}^t}, \quad \mathbf{c}_{k}^t = \frac{\bar{\mathbf{c}_{k}}^t}{||\bar{\mathbf{c}_{k}}^t||_2}
\end{equation}
where $m$ denotes the total number of received instances until the $t$-th mini-batch. Accordingly, we maintain another memory bank ($\Ocal(2N)$) to store two values for each unlabeled instance: its cluster assignment in the current epoch $a(i) = \{k | \mu_{i,k} = 1\}$ and the similarity score $s_{i, a(i)}$ (\ie the distance to its cluster centroid). 

\smallskip
\noindent \textbf{Cluster Pseudo-labels} are computed at end of each epoch by querying the memory banks. The purpose is to obtain a distribution over classes $C$ for each of our clusters based on its members. For a given cluster $k$, we obtain its label $\vz^k = [z^k_1, \cdots, z^k_C]$ as the average of the pseudo-labels of its cluster members weighted by their similarity to its centroid. Concretely, let $\Ical^k_c = \{i | \forall \vu_i \in \Ucal , a(i) = k, \hat{y}_i = c\}$ be the indices of unlabeled samples which belong to cluster $k$ and have a hard pseudo-label $c$. The probability of cluster $k$'s members belonging to class $c$ is given as:
\begin{align}
\label{eqn:clust_pl}
    z^k_c = \frac{\sum_{i \in \Ical^k_c} \quad s_{i, a(i)}}{\sum_{b=1}^C \sum_{j \in \Ical^k_b} s_{j, a(j)}}
\end{align}

\smallskip
\noindent \textbf{Refining Pseudo-labels.}
At any given epoch, we now have two pseudo-labels for an image $\vu_i$: the unrefined pseudo-label $\vp^w_i$ as well as a cluster pseudo-label $\vz^{a(i)}$ summarising its prototypical neighbourhood in the previous epoch. Accordingly, we apply our refinement procedure as follows: first, as recommended by~\cite{Berthelot_RemixMatch19, li2021comatch}, we perform distribution alignment ($DA(\cdot)$) to encourage the marginal distribution of pseudo-labels to be close to the marginal of ground-truth labels\footnote{\scriptsize $DA(\vp^w) = \vp^w / \bar{\vp^w}$, where $\bar{\vp^w}$ is a running average of $\vp^w$ during training.}, then we refine the aligned pseudo-label as per:
\vspace{-1.5mm}
\begin{align}
    \hat{\vp}^w_i = \alpha \cdot DA(\vp^w_i) + (1 - \alpha) \cdot \vz^{a(i)}
\end{align}
Here, the second term acts as a regulariser to encourage $\hat{\vp}^w$ to be similar to its cluster members' and $\alpha$ is a trade-off scalar parameter. Importantly, the refinement here leverages information based on the entire training set last-epoch information. This is in contrast to previous work~\cite{li2021comatch, zheng2022simmatch} which only stores a limited history of soft pseudo-labels for refinement, due to more memory requirement ($\Ocal(N \times C)$).

\smallskip
\noindent \textbf{Classification Loss.}
With the refined pseudo-label, 
we apply the unlabeled loss against the model prediction for the strong augmentation as per:
\vspace{-1.5mm}
\begin{equation}
\label{eqn:loss_u}
	\Lcal_u = \frac{1}{\mu B} \sum_{i=1}^{\mu B} \eta_i \cdot \mathrm{CE}(\hat{\vp}_i^w,\vp^s_i),
\end{equation}
where $\mathrm{CE}$ denotes cross-entropy.
However, unlike ~\cite{Sohn_fixmatch20, Berthelot_MixMatch19}, we do not use hard pseudo-labels or sharpening, but instead use the soft pseudo-label directly.
Also, we apply a supervised classification loss over the labeled data as per:
\vspace{-1.5mm}
\begin{equation}
	\Lcal_x = \frac{1}{B} \sum_{i=1}^{B} \mathrm{CE}(\vy_i,\vp^x_i)),
\end{equation}


\noindent \textbf{Self-supervised Loss.}
Since we use confidence as a measure of reliability (see Eqn.~\ref{eqn:loss_u}), early epochs of training suffer from limited supervisory signal when the model is not yet confident about unlabeled samples, leading to slow convergence and unstable training. Our final ingredient addresses this by introducing a consistency loss in the prototypical space on samples which fall below the confidence threshold $\tau$. We draw inspiration from instance-consistency self-supervised methods such as BYOL~\cite{grill2020bootstrap} and DINO~\cite{caron2021emerging}. In contrast to contrastive instance discrimination~\cite{he2020momentum, chen2020simple}, the former imposes consistency between two (or more) views of an image without using negative samples. Thereby, we found it to be more aligned with classification tasks than the latter. Formally, we treat the projection $q$ as soft classes score over $d$ dimensions, and obtain a distribution over these classes via a sharpened softmax ($SM(\cdot)$). We then enforce consistency between the weak and strong views as per: 
\begin{align}
\footnotesize
    \Lcal_c = \frac{1}{\mu B} \sum_{i=1}^{\mu B} (1-\eta_i) \cdot \mathrm{CE}(SM(\vq^w_i / 5T), SM(\vq^s_i / T)) 
\end{align}
Note that, as in DINO \cite{caron2021emerging}, we sharpen the target distribution more than the source's to encourage entropy minimization~\cite{Grandvalet_EntropMin05}. Unlike DINO, we do not use a separate EMA model to produce the target, we just use the output of the model for the weak augmentation. Note that this does not lead to representation collapse~\cite{grill2020bootstrap} because the network is also trained with additional semi-supervised losses.

\smallskip
\noindent \textbf{Final Objective.} We train our model using a linear combination of all four losses $\Lcal = \Lcal_x + \lambda_u \Lcal_u + \lambda_p \Lcal_p + \lambda_c \Lcal_c$. Empirically, we find that fixing $\forall \lambda = 1$, the coefficients to modulate each loss, works well across different datasets. Algorithm~\ref{alg:code} describes one epoch of \putouralg training.

\subsection{Design Considerations}
\noindent \textbf{Number of Clusters} is a crucial parameter in our approach. In essence, we refine a sample prediction obtained by the classifier by aggregating information from its $n$ nearest neighbours. However, naively doing nearest-neighbour retrieval has two limitations: 1) it requires storing image features throughout an epoch which is memory expensive; and 2) it requires a separate offline nearest-neighbour retrieval step. Instead, we leverage online clustering to identify nearest-neighbours on-the-fly. To avoid tuning $K$ for each dataset, we tuned $n$ once instead, then $K$ can be simply calculated as $K= N/n$. Additionally, we fix $\gamma=0.9n$ to ensure that each cluster contains sufficient samples to guarantee the quality of the cluster pseudo-label while relaxing clusters to not necessarily be equi-partitioned. Empirically, we found that using $n=250$ works reasonably well across datasets. To put it in context, this corresponds to $K=4800$ for ImageNet, and $K=200$ for CIFAR datasets.

\smallskip
\noindent \textbf{Multi-head Clustering} is another way to ensure robustness of our cluster pseudo-labels. To account for K-means stochastic nature, we can employ multi-head clustering to get different cluster assignments based on each head, at negligible cost. Subsequently, we can average the cluster pseudo-labels across the different heads. In practice, we find that for large datasets \eg ImageNet, cluster assignments slightly vary between heads so it is useful to use dual heads, while for smaller datasets, a single head is sufficient. 

\begin{table*}[h!]
\centering
\caption{\small{CIFAR and Mini-ImageNet accuracy for different amounts of labeled samples averaged over 5 different splits. All results are produced using the same codebase and same splits.}}
\label{tab:cifar_results}
\scalebox{0.77}{
\begin{tabular}{llccclccclcc}
\toprule[0.15em]
 &  & \multicolumn{3}{c}{\bf CIFAR-10} &  & \multicolumn{3}{c}{\bf CIFAR-100} &  & \multicolumn{2}{c}{\bf Mini-ImageNet} \\ \cline{3-5} \cline{7-9} \cline{11-12} 
Total labeled samples &  & 20 & 40 & 80 &  & 200 & 400 & 800 &  & 400 & 1000 \\ \midrule[0.15em]
FixMatch~\cite{Sohn_fixmatch20} &  & 82.32$\pm$9.77 & 86.29$\pm$4.50 & 92.06$\pm$0.88 &  & 35.37$\pm$5.68 & 51.15$\pm$1.75 & 61.32$\pm$0.92 &  & 17.18$\pm$6.22 & 39.03$\pm$3.99 \\
FixMatch + DA~\cite{Sohn_fixmatch20, Berthelot_RemixMatch19} &  & 83.84$\pm$8.35 & 86.98$\pm$3.40 & 92.29$\pm$0.86 &  & 41.28$\pm$6.03 & 52.65$\pm$2.32 & 62.12$\pm$0.79 &  & 19.40$\pm$5.87 & 40.92$\pm$4.71\\
CoMatch~\cite{li2021comatch} &  & 87.37$\pm$8.47 & 93.09$\pm$1.39 & 93.97$\pm$0.62 &  & 47.92$\pm$4.83 & 58.17$\pm$3.52 & \bf 66.15$\pm$0.71 &  & 21.29$\pm$6.19 & 40.98$\pm$3.52\\
SimMatch~\cite{zheng2022simmatch} &  & 89.31$\pm$7.73 & 94.51$\pm$2.56 & 94.89$\pm$1.32 &  & 46.01$\pm$6.12 & 57.95$\pm$2.37 & 65.50$\pm$0.93 &  & 25.75$\pm$5.90 & 39.76$\pm$3.77\\
FixMatch + DB~\cite{wang2022debiased} &  & 89.02$\pm$6.37 & 94.60 $\pm$1.31 & 95.60 $\pm$0.12 &  & 46.36$\pm$5.05 & 57.88$\pm$3.34 & 64.84$\pm$0.85 &  & 27.37$\pm$7.01 & 41.05$\pm$3.34\\
\midrule[0.15em]
\rowcolor{_fbteal3}
\putouralg  &  & \bf 90.51$\pm$4.02 & \bf 95.20$\pm$1.8 & \bf 96.11$\pm$0.20 &  & \bf 48.25$\pm$4.87 & \bf 59.53$\pm$2.94 & 65.91$\pm$0.57 &  & \bf 29.15$\pm$6.98 & \bf 45.83$\pm$4.15\\
\emph{delta against best baseline}  &  & {\color{darkgreen} +1.20} & {\color{darkgreen} +0.60} & {\color{darkgreen} +0.51} &  & {\color{darkgreen} +0.33} & {\color{darkgreen} +1.36} & {\color{darkred} -0.24} &  & {\color{darkgreen} +1.78} & {\color{darkgreen} +4.78}\\
\bottomrule[0.15em]
\end{tabular}%
}

\end{table*}



\smallskip
\noindent \textbf{Memory Analysis.} \putouralg is particulary useful due to its ability to leverage the entire prediction history in an epoch to approximate class density over neighbourhoods (represented by cluster pseudo-labels) with low memory cost. Particularly, it requires an overall of $\Ocal(4N + K \times C)$: $4N$ to store hard pseudo-labels, reliability, cluster assignments, and similarity scores; and $K \times C$ to store the cluster pseudo-labels. In contrast, if we were to employ a naive offline refinement approach, this would require $\Ocal(N \times d)$ to store the image embeddings for an epoch. For ImageNet dataset this translates to 9.6M memory units for \putouralg opposed to 153.6M for the naive approach\footnote{considering $d=128$} which is a 16$\times$ reduction in memory;  beside, eliminating the additional time needed to perform nearest neighbour retrieval.

\begin{table}[h!]
\centering
\caption{\small{DomainNet accuracy for 2, 4, and 8 labels per class.}}
\label{tab:domain_results}
\scalebox{0.7}{
\begin{tabular}{llccclccc}
\toprule[0.15em]
 &  & \multicolumn{3}{c}{\bf Clipart} &  & \multicolumn{3}{c}{\bf Sketch} \\ \cline{3-5} \cline{7-9} 
Total labeled samples &  & 690 & 1380 & 2760 &  & 690 & 1380 & 2760 \\ \midrule[0.15em]
FixMatch~\cite{Sohn_fixmatch20} &  & 30.21 & 41.21 & 51.29 &  & 12.73 & 21.65 & 33.07 \\
CoMatch~\cite{li2021comatch} &  & 35.49 & 48.62 & 54.98 &  & 24.30 & 33.71 & 41.02 \\
FixMatch + DB~\cite{wang2022debiased} &  & 38.97 & 51.44 & 58.31 &  & 25.34 & 35.58 & 43.98 \\
\midrule[0.15em]
\rowcolor{_fbteal3}
\putouralg  &  & \bf 43.72 & \bf 55.66 & \bf 61.32 &  & \bf 33.94 & \bf 43.51 & \bf 50.88 \\
\emph{delta}  &  & {\color{darkgreen} +4.75} & {\color{darkgreen} +4.22} & {\color{darkgreen} +3.01} &  & {\color{darkgreen} +8.60} & {\color{darkgreen} +7.93} & {\color{darkgreen} +6.90} \\
\bottomrule[0.15em]
\end{tabular}%
}
\vspace{-2mm}

\end{table}

\section{Experiments}
\label{experiments}

We begin by validating \putouralg's performance on multiple SSL benchmarks against state-of-the-art methods. Then, we analyse the main components of \putouralg to verify their contribution towards the overall performance, and we perform ablations on important hyperparameters.

\subsection{Experimental Settings}
\noindent \textbf{Datasets.} We evaluate \putouralg on five SSL benchmarks. Following~\cite{Sohn_fixmatch20, xie2019_uda, arazo_pseudo}, we evaluate on \textbf{CIFAR-10(100)}~\cite{cifar100} datasets, which comprises 50,000 images of 32x32 resolution of 10(100) classes; as well as the more challenging \textbf{Mini-ImageNet} dataset proposed in~\cite{mini_imagenet}, having 100 classes with 600 images per class (84x84 each). We use the same train/test split as in~\cite{label_prop} and  create splits for 4 and 10 labeled images per class to test \putouralg in the low-label regime. We also test \putouralg's performance on the \textbf{DomainNet}~\cite{peng2019moment_domainnet} dataset, which has 345 classes from six visual domains: \emph{Clipart}, \emph{Infograph}, \emph{Painting}, \emph{Quickdraw}, \emph{Real}, and \emph{Sketch}. We evaluate on the \emph{Clipart} and \emph{Sketch} domains to verify our method's efficacy in different visual domains and on imbalanced datasets.  Finally, we evaluate on \textbf{ImageNet}~\cite{russakovsky2015imagenet_large} SSL protocol as in~\cite{caron2020unsupervised, chen2020simple, caron2021emerging, assran2021semi}.
In all our experiments, we focus on the low-label regime. 

\smallskip
\noindent \textbf{Implementation Details.}
For CIFAR-10(100), we follow previous work and use WideResent-28-2(28-8)~\cite{zagoruyko2016_wideresnet} as our encoder. We use a 2-layer projection MLP with an embedding dimension $d=64$. The models are trained using SGD with a momentum of 0.9 and weight decay of 0.0005(0.001) using a batch size of 64 and $\mu=7$. We set the threshold $\tau=0.95$ and train our models for 1024 epochs for a fair comparison with the baselines. However, we note that our model needs substantially fewer epochs to converge (see Fig.~\ref{fig:analysis_plots}-b and c). We use a learning rate of 0.03 with a cosine decay schedule. We use random horizontal flips for weak augmentations and RandAugment~\cite{cubuk2020_randaugment} for strong ones. For the larger datasets: ImageNet and DomainNet, we use a Resnet-50 encoder and $d=128$, $\mu=5$ and $\tau=0.7$ and follow the same hyperparameters as in~\cite{Sohn_fixmatch20} except that we use SimCLR~\cite{chen2020simple} augmentations for the strong view. For \putouralg-specific hyperparameters, we consistently use the same parameters across all experiments: we set $n$ to 250 (corresponding to $K$=200 for CIFARs, and Mini-ImageNet, and 4800 for ImageNet), and dual learning rate $\lambda = 20$, mixing ratio $\alpha=0.8$, and temperature $T=0.1$.

\noindent \textbf{Baselines.}
Since our method bears the most resemblance with CoMatch~\cite{li2021comatch}, we compare against it in all our experiments. CoMatch uses graph contrastive learning to refine pseudo-labels but uses a memory bank to store the last n-samples embeddings to build the graph. Additionally, we compare with state-of-the-art SSL method (DebiasPL)~\cite{wang2022debiased}, which proposes a pseudo-labeling debiasing plug-in to work with various SSL methods in addition to an adaptive margin loss to account for inter-class confounding. Finally, we also compare with the seminal method FixMatch and its variant with Distribution alignment (DA). We follow Oliver~\etal~\cite{oliver_realistic} recommendations to ensure a fair comparison with the baselines, where we implement/adapt all the baselines using the same codebase to ensure using the same settings across all experiments. As for ImageNet experiments, we also compare with representation learning baselines such as SwAV~\cite{caron2020unsupervised}, DINO~\cite{caron2021emerging}, and SimCLR~\cite{chen2020simple}, where we report the results directly from the respective papers. We also include results for \putouralg and DebiasPL with additional pretraining (using MOCO~\cite{he2020momentum}) and the Exponential Moving Average Normalisation method proposed by~\cite{cai2021exponential} to match the settings used in~\cite{wang2022debiased, cai2021exponential}.

\subsection{Results and Analysis}
\noindent \textbf{Results.}
Similar to prior work, we report the results on the test sets of respective datasets by averaging the results of the last 10 epochs of training. For CIFAR and Mini-ImageNet, we report the average and standard deviation over 5 different labeled splits, whereas we report for only 1 split on larger datasets (ImageNet and DomainNet). Different from most previous work, we only focus on the very low-label regime (2, 4, and 8 samples per class, and 0.2\% for ImageNet). As shown in Tab.~\ref{tab:cifar_results} - 
\ref{tab:imagenet_results}, we observe that \putouralg outperforms baselines in almost all the cases showing a clear advantage in the low-label regime. It also exhibits less variance across the different splits (and the different runs within each split). These results suggest that besides achieving high accuracy, \putouralg shows robustness and consistency across splits in low-data regime.

Notably, our method performs particularly well on DomainNet. Unlike ImageNet and CIFARs, DomainNet is an imbalanced dataset, and prior work~\cite{tan2020class} shows that it suffers from high level of label noise. This shows that our method is also more robust to noisy labels. This can be explained in context of our co-training approach:  using the prototypical neighbourhood label to smooth the softmax label is an effective way to minimise the effect of label noise. In line with previous findings~\cite{li2021learning}, since in prototypical learning, all the instances of a given class are used to calculate a class prototype which is then used as a prediction target, it results in representations which are more robust to noisy labels.  

Finally, on ImageNet (Tab.~\ref{tab:imagenet_results}), we improve upon the closest baseline with gains of 2.2\% in the challenging 0.2\% setting; whereas we slightly fall behind PAWS~\cite{assran2021semi} in the 10\% regime, again confirming our method's usefulness in the label-scarce scenario.

\begin{figure*}[h!]
 \centering
  \scalebox{0.99}{\includegraphics[width=0.97\textwidth]{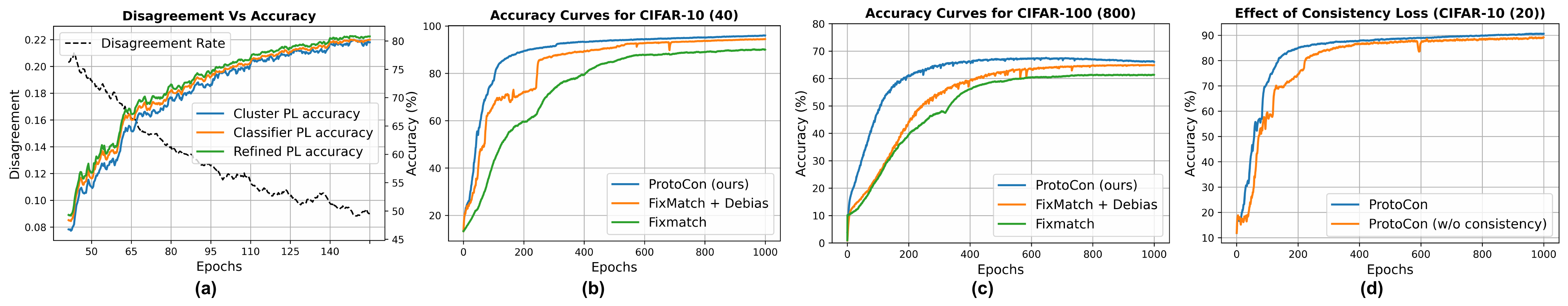}}
 \caption{\textbf{Analysis Plots.} {\bf (a)}: Average disagreement between cluster and classifier pseudo-labels versus ground truth accuracy of the different pseudo-labels. The accuracy gap between refined pseudo-labels (green) and the cluster's and classifier's (blue and orange) decreases with disagreement rate (dashed black) showing that refinement indeed helps. {\bf (b), (c):} Convergence plots on CIFAR10/100 show that \putouralg converges faster due to the additional self-supervised training signal. {\bf (d): } \putouralg w and w/out consistency loss. }
 \label{fig:analysis_plots}
 \vspace{-2mm}
\end{figure*}

\begin{figure}[h!]
 \centering
  \scalebox{0.99}{\includegraphics[width=0.47\textwidth]{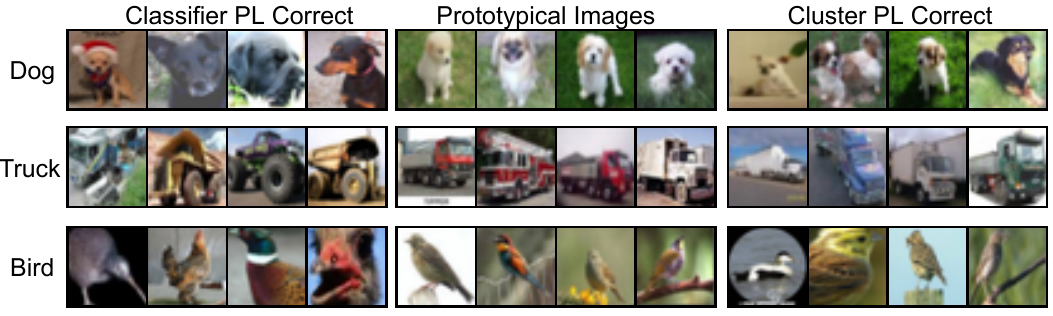}}
 \caption{The middle panel shows the most prototypical images of CIFAR-10 classes as identified by our model. Left (resp. right) panels show images which have more accurate classifier (resp. cluster) pseudo-labels. Cluster labels are more accurate for prototypical images while classifier labels are more accurate for images with distinctive features (\eg truck wheels) even if not so prototypical. Such diversity of views is key to the success of our co-training method.}
 \label{fig:image_comparison}
 \vspace{-2mm}
\end{figure}

\begin{table}[t]
\centering
\caption{\small SSL results on ImageNet with different percentage of labels. $\dagger$ denotes results produced by our codebase. Other results are reported as appearing in the cited work. }
\label{tab:imagenet_results}
\setlength{\tabcolsep}{2pt}
\scalebox{0.9}{
    \begin{tabular}{l r c c c c}
        \small Method & \small Pre. & \small Epochs & \small 0.2\% & \small 1\% & \small 10\% \\\toprule [0.15em]
        Supervised & \xmark & 300 & -- & 25.4 & 56.4\\ \midrule[0.15em]
        \multicolumn{4}{l}{\footnotesize\itshape Representation learning methods:}\\[1mm]
        SwAV~\cite{caron2020unsupervised} & \cmark & 800 & -- & 53.9 & 70.2 \\
        SimCLRv2++ ~\cite{chen2020big} & \cmark & 1200 & --  & 60.0 & 70.5 \\
        DINO~\cite{caron2021emerging} & \cmark & 300 & -- & 55.1 & 67.8 \\
        PAWS++ ~\cite{assran2021semi} & \cmark & 300 & --  & 66.5 & \textbf{75.5} \\\midrule[0.15em]
        \multicolumn{4}{l}{\footnotesize\itshape PL \& consistency methods:}\\[1mm]
        MPL~\cite{pham2021meta} & \xmark & 800 & -- & $65.3^\dagger$ & 73.9 \\
        CoMatch~\cite{li2021comatch} & \xmark & 400 & $44.3^\dagger$ & 66.0 & 73.6 \\
        FixMatch~\cite{Sohn_fixmatch20} & \xmark & 300 & -- & 51.2 & 71.5 \\
        FMatch + DA~\cite{Sohn_fixmatch20, Berthelot_RemixMatch19} & \xmark & 300 & $41.1^\dagger$ & 53.4 & $71.5^\dagger$ \\
        FMatch + EMAN~\cite{cai2021exponential} & \cmark & 850 & 43.6 & 60.9 & 72.6 \\
        FMatch + DB~\cite{wang2022debiased} & \xmark & 300 & $45.8^\dagger$ & $63.0^\dagger$ & $71.7^\dagger$ \\
        FMatch + DB + EMAN~\cite{wang2022debiased} & \cmark & 850 & 47.9 & 63.1 & $72.8^\dagger$ \\
        \midrule[0.15em]
        \rowcolor{_fbteal3}
        \putouralg & \xmark & 300 & 47.8 & 65.6 & 73.1 \\
        \rowcolor{_fbteal3}
        \putouralg + EMAN~\cite{cai2021exponential} & \cmark & 850 & \bf 50.1 & \bf 67.2 &  73.5 \\
        \emph{delta against best baseline} &  &  & {\color{darkgreen} +2.2} & {\color{darkgreen} +0.7} &  {\color{darkred} -2.0} \\
        \bottomrule[0.15em]
    \end{tabular}}
\vspace*{-1.5em}
\end{table} 
\noindent \textbf{How does refinement help?}
First, we would like to investigate the role of pseudo-labeling refinement in improving SSL performance. Intuitively, since we perform refinement by combining pseudo-labels from two different sources (the classifier predictions in probability space and the cluster labels in the prototypical space), we expect that there will be disagreements between the two and hence considering both the views is the key towards the improved performance. To validate such intuition, we capture a fine-grained view of the training dynamics throughout the first 300 epochs of CIFAR-10 with 40 labeled instances scenario, including: samples' pseudo-labels before and after refinement as well as their cluster pseudo-labels in each epoch. This enables us to capture disagreements between the two pseudo-label sources up to the individual sample level.
In Fig.~\ref{fig:analysis_plots}-a, we display the average disagreement between the two sources over the initial phase of the training overlaid with the classifier, cluster and refined pseudo-label accuracy. We observe that initially, the disagreement (dashed black line) is high which corresponds to a larger gap between the accuracies of both heads. As the training proceeds, we observe that disagreement decreases leading to a respective decrease in the gap. Additionally, we witness that the refined accuracy curve (green) is almost always above the individual accuracies (orange and blue) which proves that, indeed, the synergy between the two sources improves the performance. 

On the other hand, to get a qualitative understanding of where each of the pseudo-labeling sources helps, we dig deeper to classes and individual samples level where we investigate which classes/samples are the most disagreed-upon (on average) throughout the training. In Fig.~\ref{fig:image_comparison}, we display the most prototypical examples of a given class (middle) as identified by the prototypical scores obtained in the embedding space. We also display the examples which on average are always correctly classified in the prototypical space (right) opposed to those in the classifier space (left). As expected, we find that samples which look more prototypical, albeit with less distinctive features (\eg blurry), are the ones almost always correctly classified with the prototypical head; whereas, samples which have more distinctive features but are less prototypical are those correctly classified by the discriminative classifier head. This again confirms our intuitions about how co-training based on both sources helps to refine the pseudo-label.
	
Finally, we ask: is it beneficial to use the entire dataset pseudo-label history to perform refinement or is it sufficient to just use a few samples? To answer this question, we use only a subset of the samples in each cluster (sampled uniformly at random) to calculate cluster pseudo-labels in Eqn.~\ref{eqn:clust_pl}. For CIFAR-10 with 20 and 40 labels, we find that this leads to about 1-2\% (4-5\%) average drop in performance, if we use half (quarter) of the samples in each cluster.  This reiterates the usefulness of our approach to leverage the history of all samples (at a lower cost) opposed to a limited history of samples. 

\smallskip
\noindent \textbf{Role of self-supervised loss.}
Here, we are interested to tear apart our choice of self-supervised loss and its role towards the performance. To recap, our intuition behind using that loss is to boost the learning signal in the initial phase of the training when the model is still not confident enough to retain samples for pseudo-labeling. As we see in Fig.~\ref{fig:analysis_plots}-b and c. there is a significant speed up of our model's convergence compared to baseline methods with a clear boost in the initial epochs. Additionally, to isolate the potentially confounding effect of our other ingredients, we display in Fig.~\ref{fig:analysis_plots}-d the performance of our method with and without the self-supervised loss which leads to a similar conclusion. Finally, to validate our hypothesis that instance-consistency loss is more useful than instance-discrimination, we run a version of \putouralg with an instance-discrimination loss akin to that of SimCLR. This version completely collapsed and did not converge at all. We attribute this to: 1) as verified by SimCLR authors, such methods work best with large batch sizes to ensure enough negative examples are accounted for; and 2) these methods treat each image as its own class and contrast it against every other image and hence are in direct contradiction with the image classification task; whereas instance-consistency losses only ensure that the representations learnt are invariant to common factors of variations such as: color distortions, orientation, \etc. and are hence more suitable for semi-supervised image classification tasks.

\begin{table}[t]
\caption{Ablation results. -$\Lcal_*$ denotes that the respective loss is not applied, and {\color{darkgreen} \bf \underline {green}} marks the best option. Results are average accuracy over 5 runs for CIFAR-10 (80).}
\label{tab:ablation}
\tiny
\centering
\setlength{\tabcolsep}{2pt}
\resizebox{0.4\textwidth}{!}{%
\scalebox{0.2}{ 
\begin{tabular}{llcccc}
\toprule[0.15em]
\multirow{2}{*}{Losses} &  & -{$\Lcal_c$} & -{$\Lcal_p$} & -{($\Lcal_c$,$\Lcal_p$)} & {\color{darkgreen} \bf \underline{All}} \\
 &  & 95.3 & 94.8 & 92.3 & 96.1 \\
\midrule[0.15em]
\multirow{2}{*}{Cluster size ($n$)} &  & 50 & {\color{darkgreen} \bf \underline{250}} & 500 & 1000 \\
 &  & 95.7  & 96.1 & 94.3 & 92.1 \\
\midrule[0.15em]
\multirow{2}{*}{Refinement Ratio ($\alpha$)} &  & 0.5 & 0.7 & {\color{darkgreen} \bf \underline{0.8}} & 0.9 \\
 &  & 86.7 & 94.5 & 96.1 & 95.2 \\
\bottomrule[0.15em]
\end{tabular}%
}}
\end{table}


\smallskip
\noindent \textbf{Ablations.}
Finally, we present an ablation study about the important hyperparametes of \putouralg. Specifically, we find that $n$ (minimum samples in each cluster) and $\alpha$ (mixing ratio between classifier pseudo-label and cluster pseudo-label) are particularly important. Additionally, we find that the projection dimension needs to be sufficiently large for larger datasets (we use $d=64$ for CIFARs and 128 for all others). In Tab.~\ref{tab:ablation}, we present ablation results on CIFAR-10 with 80 labeled instances.

\section{Conclusion}

We introduced \putouralg, a novel SSL learning approach targeted at the low-label regime. Our approach combines co-training, clustering and prototypical learning to improve pseudo-labels accuracy. We demonstrate that our method leads to significant gains on multiple SSL benchmarks and better convergence properties. We hope that our work helps to commodify deep learning in domains where human annotations are expensive to obtain. 

\smallskip
\noindent \textbf{Acknowledgement.} This work was partly supported by DARPA’s Learning with Less Labeling (LwLL) program under agreement FA8750-19-2-0501. I. Nassar is supported by the Australian Government Research Training Program (RTP) Scholarship, and M. Hayat is supported by the ARC DECRA Fellowship DE200101100.

{\small
\bibliographystyle{ieee_fullname}
\bibliography{references.bib}
}
\newpage


\setcounter{section}{0}
\setcounter{equation}{0}
\renewcommand{\thesection}{\Alph{section}}
\newpage
\newpage
\section*{Appendix}


\section{Constrained K-means Additional Details}
Qian \etal~\cite{qian2022unsupervised} proposed the online mini-batch solver for the constrained K-means objective (Eqn.~\ref{eqn:ckmeans}) proposed by~\cite{bradley2000constrained}, and used it for unsupervised representation learning. In our method, we adopted the same solver but for a different purpose; we use online clustering as an alternative to offline nearest neighbour search to identify neighbourhood of images and leverage such information to perform our label refinement procedure. To that end, due to the empirical observation that the maximal value of dual variables is well bounded, our Eqn.~\ref{eq:finalrho} is an approximation of the original dual variables update proposed by Qian \etal after each mini-batch:
\begin{eqnarray}
\label{eq:update}
\rho^t_k = \Pi_{\Delta_\delta}(\rho^{t-1}_k - \eta \frac{1}{B}\sum_{i=1}^B( \mu_{i,k}^t-\frac{\gamma}{N})),
\end{eqnarray}
where $\Pi_{\Delta_\delta}$ projects the dual variables to the domain $\Delta_\delta = \{\rho| \forall k, \rho_k\geq0, \|\rho\|_1\leq \delta\}$.

We refer the readers to the original paper for guarantees of performance complete proofs.

\begin{figure*}[h!]
 \centering
  \scalebox{0.99}{\includegraphics[width=0.97\textwidth]{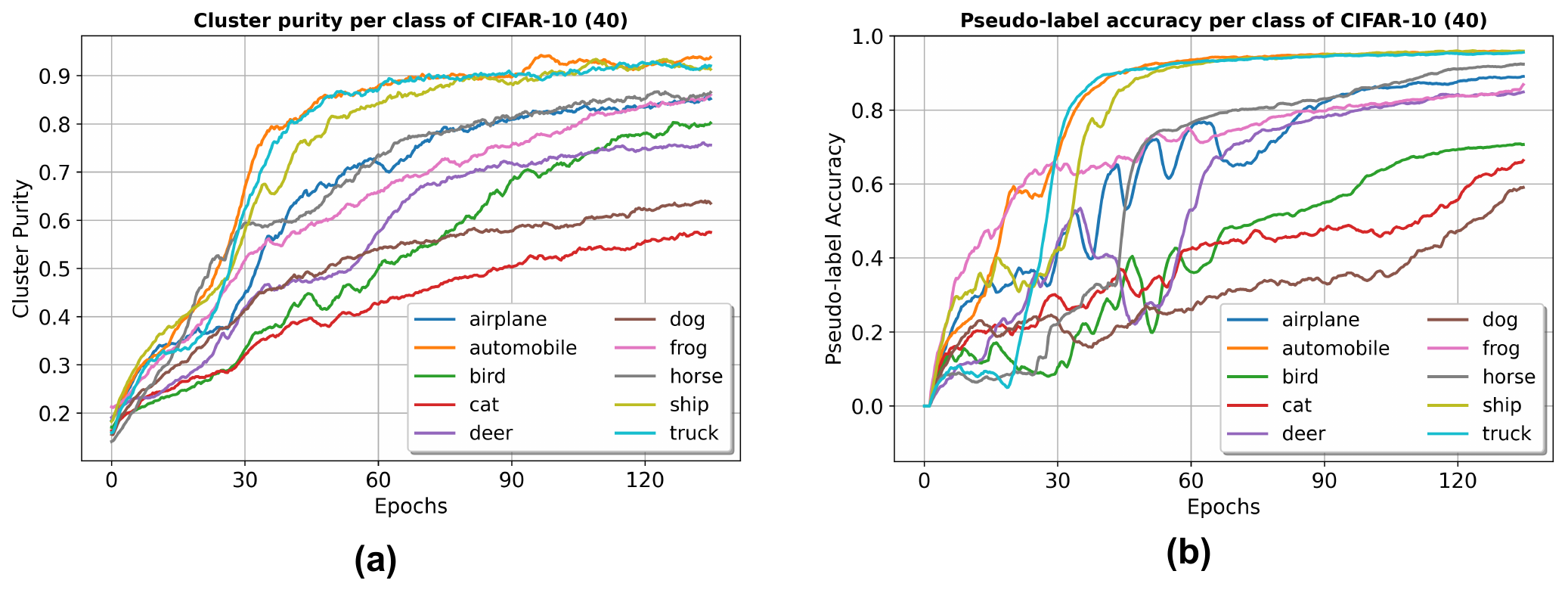}}
 \caption{\textbf{Analysis Plots.} {\bf (a)}: Cluster Purity per class of CIFAR-10 vs training epochs, when trained using \putouralg with 4 images per class. {\bf (b):} Pseudo-label accuracy per class vs training epochs. Best viewed in color.}
 \label{fig:purity_vs_acc}
 \vspace{-2mm}
\end{figure*}

\smallskip
\noindent \textbf{Constrained vs unconstrained clustering.}
Our purpose in \putouralg is to use K-means as an alternative for offline nearest-neighbours retrieval, which automatically mandates that we use equi-partition clustering by constraining minimum cluster size $\gamma$ to be the number of nearest neighbours $n$. However, we relax this constraint to $\gamma = 0.9n$ to allow cluster sizes to slightly vary to capture the inherent imbalance in salient properties of different classes. Empirically, we found this to work well across the datasets we used. We also tested the setting with $\gamma = 0$ which translates to unconstrained clustering. This setting was unstable and did not lead to performance gains; where we found that clustering collapses to only a few clusters. For example in CIFAR-10 (40 labels) setting, K-means converged to only 20 clusters. The consequence is that we have only 20 cluster pseudo-labels to use for refining all the unlabeled samples in subsequent epochs which is a very general summary of neighbourhoods and hence it hurts the performance rather than help it. Please refer to Tab.~\ref{tab:ablation} for further ablations on the value of $n$.

\smallskip
\noindent \textbf{Mini-batch updates vs Epoch updates}
Another decision choice is the frequency of cluster centroids updates (Eqn.~\ref{eq:updatemulti}). Since \putouralg does not memorise image representations, centroids can be updated either every mini-batch, or by accumulating representations of images based on their cluster assignment throughout an epoch and then performing the update once at the end of the epoch. The former solution is useful in helping K-means convergence which requires multiple assignment-update iterations, however it leads to higher variance due to the stochastic nature of mini-batches. On the other hand, the latter solution is also sub-optimal as it requires long time for clusters to converge. Accordingly, we adopted a warmup period during which we use mini-batch updates to speed up convergence, henceforward, we switch to epoch updates to stabilise the centroids and exhibit less variance. We found that for smaller datasets, 20 epochs of warmup are sufficient, while for the larger datasets with more classes, we increase the warmup period to 70 epochs. 

\section{Additional Training Dynamics Analysis}
Here, to further understand \putouralg, we examine more of its training dynamics.

\begin{figure*}[h!]
 \centering
  \scalebox{0.99}{\includegraphics[width=0.97\textwidth]{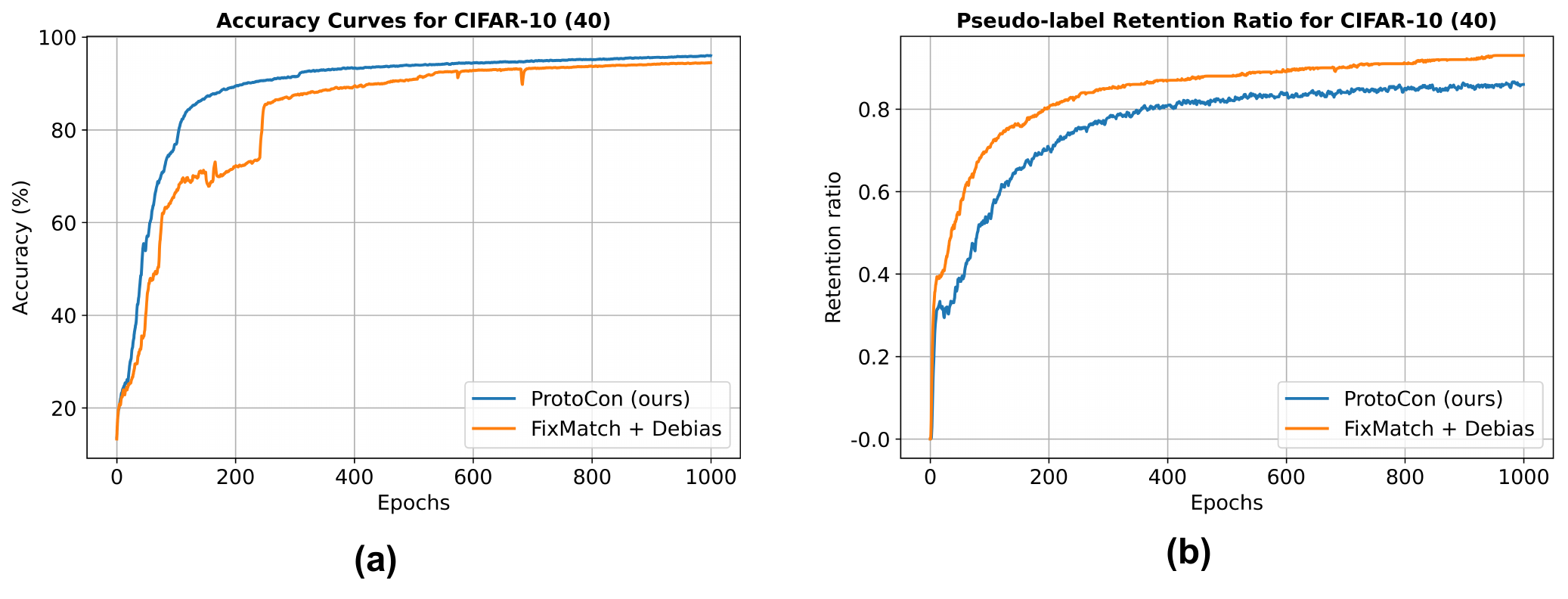}}
 \caption{\textbf{Analysis Plots.} {\bf (a)}: Pseudo-label accuracy vs epochs. {\bf (b):} Retention rate vs epochs which denotes the ratio of unlabeled samples retained by each method for pseudo-labeling (\ie with maximum confidence score higher than the threshold $\tau$.)}
 \label{fig:acc_vs_retention}
 \vspace{-2mm}
\end{figure*}

\smallskip
\noindent \textbf{Clustering purity vs pseudo-label accuracy.}
First, we investigate the properties of the clusters as training proceeds. We follow a similar setup like that used to obtain Fig.~\ref{fig:image_comparison}, but this time, we use the captured statistics to calculate cluster purity for each class. Specifically, by the end of each epoch, we count the members of each cluster (\eg for CIFAR-10, we use $K=250$, so we count the number of images assigned by K-means to each of the 250 clusters), then for each cluster, we check the most dominant class among its members based on their ground truth labels. Subsequently, we calculate the purity of each cluster as the ratio between the number of images belonging to the dominant class to the total number of cluster members. Finally, to calculate purity for a given class, we average the described ratio across all clusters for which that class is the dominant one. In Fig.~\ref{fig:purity_vs_acc}, we display cluster purity per class of CIFAR-10 during the first 130 epochs of training side-by-side to the pseudo-label accuracy for each class. This is to allow us to investigate the clustering effectivness in the critical initial phase of training and how it affects the obtained pseudo-labels quality. We see that for the more distinguishable classes (\eg truck or ship), clustering purity increases significantly faster than others matching with a corresponding increase in pseudo-label accuracy. Whereas for more confusing classes (\eg horse and deer), the cluster purity suffers a slow increase accompanied with what seem to be high disagreement between cluster and classifier pseudo-labels leading to an overall slow increase of pseudo-label accuracy (note that we display the refined pseudo-label accuracy in the figure). Finally, the most confusing classes (\eg cat and dog) have the lowest cluster purity leading to a low pseudo-label accuracy at first, but we notice that once the majority of other classes are learnt (\ie have higher accuracy, the more confusing classes start to catch up (notice the cat and dog curves towards the end of Fig.~\ref{fig:purity_vs_acc}-b). This is in line with our expectation that easy classes are first learnt by the network, then it moves on to discriminate the less obvious ones.   

\smallskip
\noindent \textbf{Pseudo-label Retention Ratio.} Like the state-of-the-art SSL method (DebiasPL~\cite{wang2022debiased}), \putouralg is also a confidence-based pseudo-labeling method albeit with additional ingredients. Hence, both methods only retain high-confidence unlabeled samples for pseudo-labeling. In Fig.~\ref{fig:acc_vs_retention}, we examine the retention rate (\ie ratio of samples with maximum confidence exceeding the threshold $\tau$) for both methods as training proceeds (b), and compare it with the pseudo-labeling accuracy exhibited by each (a). We observe that even though our method outperforms DebiasPL, in terms of accuracy, throughout the training, it consistently retains almost 10\% less samples for pseudo-labeling. This finding speaks to our original motivation (see Sec.~\ref{sec:introduction}) with regards to the over-confidence problem underpinning the lower performance of SOTA methods in label-scarce regime. Compared to its counterparts, \putouralg is more conservative when it comes to admitting a sample as ``reliable'' for pseudo-labeling; primarily because the refined pseudo-labels we employ is a combination of the original classifier pseudo-label and the neighbourhood pseudo-label. As we show in Fig.~\ref{fig:analysis_plots}-a, the disagreement between the two results in a lower overall confidence in predictions. Such conservative nature of \putouralg is key to avoiding confirmation bias even when there is only a few labeled samples available.

\begin{table*}[h!]
\centering
\caption{\small{CIFAR and Mini-ImageNet accuracy in moderate-label regime for different amounts of labeled samples averaged over 3 different splits. All results are produced using the same codebase and same splits.}}
\label{tab:cifar_results_moderate}
\scalebox{0.99}{
\begin{tabular}{llcclcc}
\toprule[0.15em]
 &  & \multicolumn{2}{c}{\bf CIFAR-100} &  & \multicolumn{2}{c}{\bf Mini-ImageNet} \\ \cline{3-4} \cline{6-7} 
Total labeled samples &  & 2500 & 4000  &  & 2500 & 4000 \\ \midrule[0.15em]
FixMatch~\cite{Sohn_fixmatch20} &  & 71.71$\pm$0.35 & 74.08$\pm$0.13 &  & 44.53$\pm$0.44 & 50.21$\pm$0.09 \\
FixMatch + DB~\cite{wang2022debiased} &  & 72.44$\pm$0.15 & 74.43$\pm$0.06 &  & 46.18$\pm$0.23 & 52.00$\pm$0.04 \\
\midrule[0.15em]
\rowcolor{_fbteal3}
\putouralg  &  & \bf 73.31$\pm$0.43 & \bf 75.18$\pm$0.02 &  & \bf 48.61$\pm$0.34 & \bf 53.67$\pm$0.06 \\
\emph{delta against best baseline}  &  & {\color{darkgreen} +0.87} & {\color{darkgreen} +0.75} &  & {\color{darkgreen} +2.43} & {\color{darkgreen} +1.67} \\
\bottomrule[0.15em]
\end{tabular}%
}

\end{table*}

\begin{figure*}[h!]
 \centering
  \scalebox{1}{\includegraphics[width=0.99\textwidth]{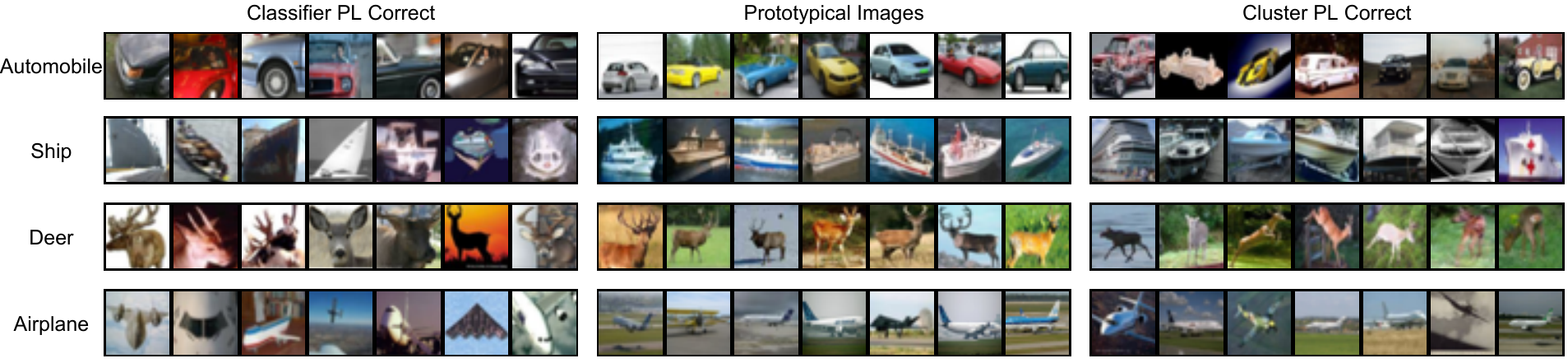}}
 \caption{Additional examples to complement Fig.~\ref{fig:image_comparison}.}
 \label{fig:image_comparison_appendix}
 \vspace{-2mm}
\end{figure*}

\section{\textbf{\putouralg} in Moderate-label Regime}
In this section, we examine our method performance when more than 10 images per class are available (which we call moderate-label regime). To recap, our method primarily aims to address confirmation bias in label-scarce settings. Yet, intuitively, the refinement strategy might also help moderate-label regimes. As such, we investigate this hypothesis by running additional experiments on CIFARs and Mini-ImageNet with 25, and 40 images per class. We find that for CIFAR-10, performance already saturates after 10 images per class and most of the compared methods perform similarly. As for the other two datasets with 100 classes each, we find \putouralg to still provide performance gains. However, with more labels available, we find that using less neighbouring samples to perform the refinement (\ie less $n$) works better. Specifically, we reduce $n$ by a factor of 10 (\ie $n=25$ instead of $n=250$). Additionally, since with more labels, all the compared methods exhibit significantly less variance, we report results only based on 3 runs instead of 5. Please refer to the results in Tab.~\ref{tab:cifar_results_moderate}.

\section{Additional Quantitative Examples}
Here, we detail our experimental setup for obtaining Fig.~\ref{fig:image_comparison} and we provide additional examples in Fig.~\ref{fig:image_comparison_appendix}.

\smallskip
\noindent \textbf{Experimental Setup.} As training proceeds, for each epoch, we capture per-image statistics such as: the classifier pseudo-label and its max score (\ie $\argmax \vp_w$ and $\max \vp_w$ respectively); cluster pseudo-label and its max score (\ie $\argmax \vz^a$ and $\max \vz^a$ respectively), sample prototypical score (\ie $\vq^w \cdot \Pcal_{\hat{y}}$) denoting how close a sample is to its class prototype. Subsequently, to obtain the prototypical images (in middle panel of Fig.~\ref{fig:image_comparison} and ~\ref{fig:image_comparison_appendix}), we rank images of each class based on their prototypical score averaged over the first 500 epochs of training. Additionally, we identify images for which the cluster pseudo-labels are, on average, more accurate than that of the classifier (and the other way around) by comparing the respective pseudo-labels with the ground truth label of each image. Thus, we display on the left panel images for which the classifier pseudo-label is, on average, more accurate than the cluster pseudo-label, and the opposite on the right panel.

\smallskip
\noindent \textbf{Additional Examples.} In Fig.~\ref{fig:image_comparison_appendix}, we provide more examples to complement those in Fig.~\ref{fig:image_comparison}. To reiterate, we see that the cluster pseudo-labels which capture the samples' neighbourhood in the prototypical space (trained via our prototypical loss) are usually more accurate if images are more prototypical even if they are lacking discriminative features (\eg blurry images or zoomed out images). In contrast, the pseudo-labels in the class probability space (trained via one-hot cross entropy) are usually more accurate for images with discriminative features (\eg car bumpers or deer horns) even if they lack prototypicality. The diversity of views captured via the different labels is key to \putouralg's effectiveness as it helps the classifier learns via the disagreement between the two views through the refined label.
\end{document}